\pdfoutput=1
\documentclass[twoside,11pt]{article}

\usepackage{dmlr2e}

\usepackage{hyperref}       
\usepackage{url}            
\usepackage{booktabs}       
\usepackage{amsfonts}       
\usepackage{nicefrac}       
\usepackage{microtype}      
\usepackage[ruled]{algorithm2e}
\usepackage{threeparttable}

\usepackage{graphicx}
\usepackage{subfigure}
\usepackage{caption}
\usepackage{subcaption}
\usepackage{booktabs} 
\usepackage{framed}
\usepackage{amssymb}
\usepackage{mathrsfs}
\usepackage{mathtools}
\usepackage{array}
\usepackage{verbatim} 
\usepackage{enumerate}
\usepackage{bbm}
\usepackage{commath}
\usepackage{wrapfig}
\usepackage{amsbsy}
\usepackage{float}
\usepackage{soul}
\usepackage{xcolor}
\usepackage{amsmath}
\usepackage{tabularx}
\usepackage{listings}
\usepackage{colortbl}
\usepackage[shortlabels]{enumitem}
\usepackage{tcolorbox}
\usepackage{multirow}
\usepackage{graphicx}
\usepackage{tcolorbox}
\usepackage{paralist}

\DeclareMathOperator*{\argmax}{arg\,max}

\definecolor{comment}{RGB}{70, 150, 60}

\newenvironment{myitemize}{%
\begin{itemize}[leftmargin=1em, itemsep=.1em, parsep=.1em, topsep=.1em,
    partopsep=.1em]}
{\end{itemize}}

\newenvironment{myenumerate}{%
\begin{enumerate}[leftmargin=1em, itemsep=.1em, parsep=.1em, topsep=.1em,
    partopsep=.1em]}
{\end{enumerate}}

\newenvironment{structure*}{\color{blue}\begin{myenumerate}}{\end{myenumerate}}

\hypersetup{
    colorlinks,
    linkcolor={red!50!black},
    citecolor={blue!50!black},
    urlcolor={blue!80!black}
}

\definecolor{lightorange}{RGB}{255,229,204}
\sethlcolor{lightorange}

\definecolor{lightblue}{RGB}{173,216,230}

\newcommand{\method}{\textsc{Synthetic}\xspace}

\usepackage{lastpage}
\dmlrheading{24}{2024}{1-\pageref{LastPage}}{5/31; Revised 07/12}{06/17}{21-0000}{Light, Liu, and Hu} 
\def\openreview{\url{https://openreview.net/forum?id=yf11dMbvMV}} 

\ShortHeadings{Dataset Distillation for Offline Reinforcement Learning}{Light, Liu, and Hu}
\firstpageno{1}

\begin{document}
\title{Dataset Distillation for Offline Reinforcement Learning}

\author{\name Jonathan Light$^*$ \email lij@rpi.edu \\
       \addr Department of Computer Science \\
       Rensselaer Polytechnic Institute \\
       Troy, NY 12180, USA
       \AND
       \name Yuanzhe Liu$^*$ \email liuy72@rpi.edu \\
       \addr Department of Computer Science \\
       Rensselaer Polytechnic Institute \\
       Troy, NY 12180, USA
       \AND
       \name Ziniu Hu \email acgbull@gmail.com \\ 
       \addr Computing + Mathematical Sciences Department \\
       California Institute of Technology\\
       Los Angeles, CA 91125, USA}

%



\editor{}
\maketitle

\def\thefootnote{*}\footnotetext{These authors contributed equally to this work}\def\thefootnote{\arabic{footnote}}

\begin{abstract}
    Offline reinforcement learning often requires a quality dataset that we can train a policy on. However, in many situations, it is not possible to get such a dataset, nor is it easy to train a policy to perform well in the actual environment given the offline data. 
    We propose using data distillation to train and distill a better dataset which can then be used for training a better policy model.
    We show that our method is able to synthesize a dataset where a model trained on it achieves similar performance to a model trained on the full dataset or a model trained using percentile behavioral cloning. Our project site is available \href{https://datasetdistillation4rl.github.io}{here}. We also provide our implementation at \href{https://github.com/ggflow123/DDRL}{this GitHub repository}.
\end{abstract}

\section{Introduction}

A significant challenge in reinforcement learning (RL) is that the data generation process is coupled with the training process, and data generation requires frequent online interaction with the environment, which is not possible in many settings.
Offline RL aims to solve this problem by decoupling the two and training the agent on a given a static, fixed dataset (\cite{levine2020offline},\cite{prudencio2023survey}).
However, offline RL relies on a dataset generated by a good expert policy. We often do not have access to data generated by good policies, only mediocre ones. Offline training also means we face the distributional shift problem, where the policy trained on the dataset is produces a different data distribution than the one in the dataset.

Instead of taking the usual offline RL approach of finding a better way to train a model given the offline dataset, we take an alternate approach of asking, is there a way to distill a better offline dataset to train on? We believe that this approaches offers several advantages over finding a better training method. First of all, it is easier to \emph{interpret} a distilled dataset vs a better trained model. Secondly, distillation tends to lead to better generalization capabilities since we learn the key features of the input space (\cite{ stanton2021does}, \cite{sachdeva2023data}). Thirdly, a distilled dataset is much smaller than the original offline dataset, which improves sample efficiency.

We propose using a method from data distillation  \cite{wang2018dataset} known as gradient matching to train a smaller synthetic dataset on the offline dataset. 
We evaluated the students trained using our procedure on the Procgen environment \cite{pmlr-v119-cobbe20a}, which consists of procedurally generated games. 
Specifically, the student is only given access to offline expert policy data on some of the procedurally generated maps, and must generalize the knowledge they learn on those maps to other unseen, out-of-distribution settings. 
We show that students trained using synthetic data are able to perform similarly or better than students trained on the original offline policy dataset or students trained using percentile behavorial cloning both in distribution and out of distribution, despite the fact that they are \emph{trained on a smaller dataset.}

Why does training on a smaller dataset help in RL settings specifically? 
RL is a learning paradigm that is natural very prone to randomness and over-fitting due to the fact that the agent also has control of the data generation process. 
The insight that we have here is that a smaller and well controlled dataset can reduce randomness and overfitting. 
Just like how humans learn more effectively when read a well written book instead of reading many low quality articles, reinforcement learning agents can also learn a better, more generalizable policy by training on a high quality dataset. 



To summarize, our main contributions are as follows:
\begin{compactitem}
    \item We propose a new method \method that synthesizes a new dataset given a offline dataset of trajectories generated by an expert policy using dataset distillation. 
    \item We show that a RL-model trained on a dataset synthesized using our method is able to perform similarly or better in the environment than directly training on the expert data or other techniques such as percentile behavioral cloning 
    \item We demonstrate how we are able to achieve similar performance with a far smaller dataset when training the RL-model on our synthesized dataset 
\end{compactitem}



\section{Methodology}
We describe our general problem setting, the baseline methods of tackling the problem, and our method here. 
\subsection{Offline reinforcement learning problem setting}
\label{sec:offline-rl}
In our reinforcement learning setting, the environment is modelled as a \textbf{Markov decision process (MDP)} $\mathcal{M} = \langle \mathcal{S}, \mathcal{A}, T, R, s_0, s_{-1}\rangle$ with state space $\mathcal{S}$, action space $\mathcal{A}$, $T(s_{t+1} | s_t, a)$ is the probabilistic transition function, $R(s_t, a) \in \mathbb{R}$ is the immediate transition reward, $s_0$ is the start state, and $s_{-1}$ is the end state since we are in an episodic setting. A policy function $\pi: \mathcal{S} \rightarrow \Delta (\mathcal{A})$ is a mapping from a state to a probability distribution over the action space. We will assume that both the $\mathcal{S}$ and $\mathcal{A}$ are discrete in our setting. When a policy can be parameterized by some parameter $\theta$, we denote the policy as $\pi_\theta$. Since we are using deep reinforcement learning, $\pi_\theta$ will be a neural network with weights $\theta$.

The goal of parametrized reinforcement learning is to learn the optimal $\theta^*$ that maximizes the cumulative of an episode. 
We define the cumulative reward in terms of the trajectory distribution induced by the policy $\pi_\theta$. A trajectory $\tau$ is a sequence of states and actions that starts with $s_0$ and ends with $s_{-1}$, i.e. $\tau = ((s_0, a_0), (s_1, a_1), ..., (s_{-1}, a_{-1}))$. Then the trajectory distribution $p_{\pi}$ induced by policy $\pi$ and environment $\mathcal{M}$ is given by 
\[p_\pi (\tau) = \prod_{(s_t,a_t) \in \tau} \pi({a_t | s_t}) T(s_{t+1} | s_t, a_t) \]
The expected cumulative reward is then
\[J(\pi) = \mathbb{E}_{\tau \sim p_\pi} \left[ \sum_{(s,a) \in \tau} R(s,a)\right]\]
and our goal is to find
\[\theta^* = \argmax_{\theta} J(\pi_\theta), \quad \max_{\theta} J(\pi_\theta) \approx \max_\pi J(\pi) \]

In offline reinforcement learning, the agent is not allowed to interact with the environment and collect data through that. Instead, we are given a static data set $\mathcal{D} = \{(s_t^i, a_t^i, s_{t+1}^i, r_t^i)\}$ of transitions to learn the best policy $\pi_\theta$ from, where $(s_t^i, a_t^i) \in \tau^i$ are part of the trajectory of episode $i$. Sometimes the dataset also includes the future return $G_t^i = \sum_{(s_k^i, a_k^i)\in \tau^i, k \ge t} R(s,a)$ at both the current state and for the whole episode, so $\mathcal{D} = \{(s_t^i, a_t^i, s_{t+1}^i, r_t^i, G_t^i, G_0^i)\}$. We assume that the data is generated using some policy $\pi_\beta$, i.e. $\tau^i \sim p_{\pi_\beta}$. 

\subsection{Behavioral cloning}
\label{sec:behavioral_cloning}
A common way to approach offline RL is to simply attempt to train a policy $\pi_\theta$ to imitate $\pi_{\beta}$. This is done in a supervised learning fashion by training $\pi_\theta$ to predict $a_t^i$ given $s_t^i$ and minimizing the loss 
\begin{equation}
\label{eq:bc_loss}
    \mathcal{L}_{\textbf{BC}}(\theta | \mathcal{D}) = \sum_{(s_t^i, a_t^i, ...) \in \mathcal{D}} w(s_t^i, a_t^i, ...) ||\pi_\theta(s_t^i) - a_t^i|| 
\end{equation} 
where $||$ is some norm implying some notion of distance and $w$ is some weighting on the data-points in $\mathcal{D}$. Usually we take the uniform weighting $w(*) = 1$. Since we usually use stochastic gradient descent (SGD) to minimize the loss, in this case we can simply use $p(*) = \frac{w(*)}{\sum_{(s_t^i, a_t^i, ...) \in \mathcal{D}} w(s_t^i, a_t^i, ...)}$ as the probability of selecting the sample. 

Since $\pi_\beta$ might not be optimal, we can attempt to train a better policy $\pi_\theta$ by filtering out observations in $\mathcal{D}$ that lead to poor outcomes. In other words, we let $w(s_t^i, a_t^i, ..., G_0^i) = \mathbb{I}_{G_0^i \ge b} $ where $b$ is some threshold on good vs bad outcomes. We call the method of choosing $b$ such that we only end up with x\% of the initial dataset, and training a policy $\pi_\theta$ using BC on it as BCx\%, or \textbf{percentile behavior cloning}. In other words, the goal of  percentile behavior cloning is to filter out a better training dataset.


\subsection{Synthetic dataset}
Instead of filtering out bad samples in order to create a good training dataset, our method \emph{directly learns a good training sample} instead. 
This is done through dataset distillation, a technique used in supervised learning \cite{wang2018dataset}. 
We described how we `train' a synthesized dataset $\mathcal{D}_\phi$ here, parameterized by $\phi$, given the offline dataset $\mathcal{D}_\text{real}$.
Our method aims to reduce the gradient matching loss of $\phi$ with respect to some random initialization of the model weights $\theta$ according to some distribution $\theta \sim p_\theta$.
Given some model parameters $\theta$, we first get the gradient of $\theta$ with respect to the BC loss we defined in \ref{eq:bc_loss} on both the real dataset, $\nabla_\theta \mathcal{L}_{\text{BC}} (\theta |  \mathcal{D}_{\text{real}})$, and our synthetic one, $\nabla_\theta \mathcal{L}_{\text{BC}} (\theta| \mathcal{D}_\phi)$. 
Then we define the gradient matching loss as 
\begin{equation}
    \label{eq:gradient_matching_loss}
    \mathcal{L}_{\text{grad match}}(\phi | \theta_i) = \mathbb{E}_{\theta \sim p_\theta} \left[ ||\nabla_\theta \mathcal{L}_{\text{BC}} (\theta| \mathcal{D}_{\text{real}}) - \nabla_\theta \mathcal{L}_{\text{BC}} (\theta| \mathcal{D}_\phi)|| \right]
\end{equation}
We then use SGD to minimize this loss. 
This method helps guarantee that the synthesized dataset $\mathcal{D}_\text{syn}$ will produce a gradient similar to that of $\mathcal{D}$ when a model is trained on it. 

\begin{figure}[hbt!]
    \centering
    \includegraphics[width=1.0\linewidth]{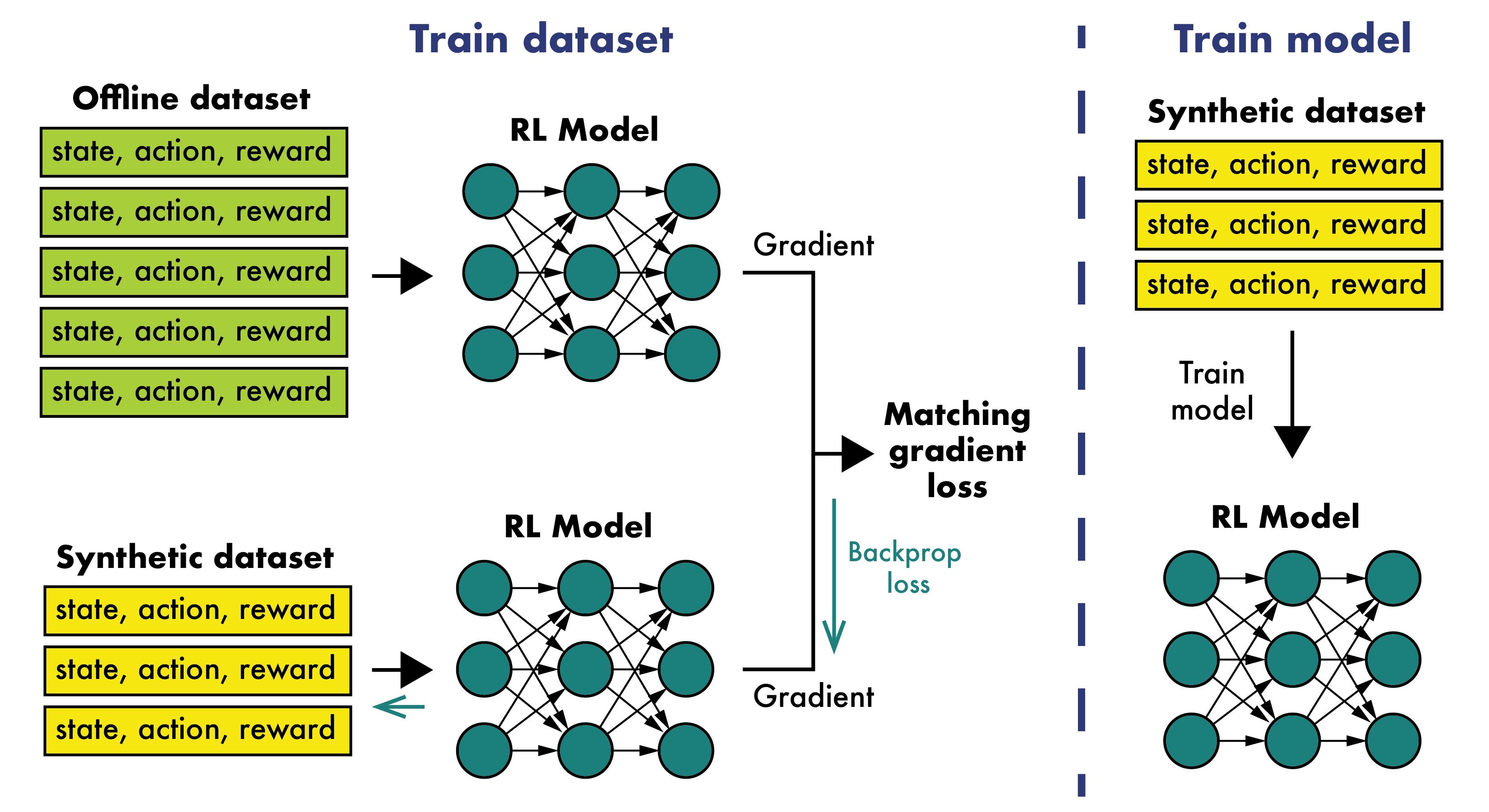}
    \caption{\textbf{Overview of our dataset distillation process.} On the left we train the dataset by taking the matching gradient loss between the real offline dataset and our synthetic dataset. On the right we then use the trained synthetic dataset to train a RL model, which we then evaluate on the real environment.}
    \label{fig:offline-distillation}
\end{figure}

\section{Experimental Setup}

We provide details on how we implemented our experiments below, including what environments we tested our method on, the architecture for our models, and how we trained the models.

\subsection{Environment}
\begin{figure}[hbt!]
    \centering
    \includegraphics[width=0.33\linewidth]{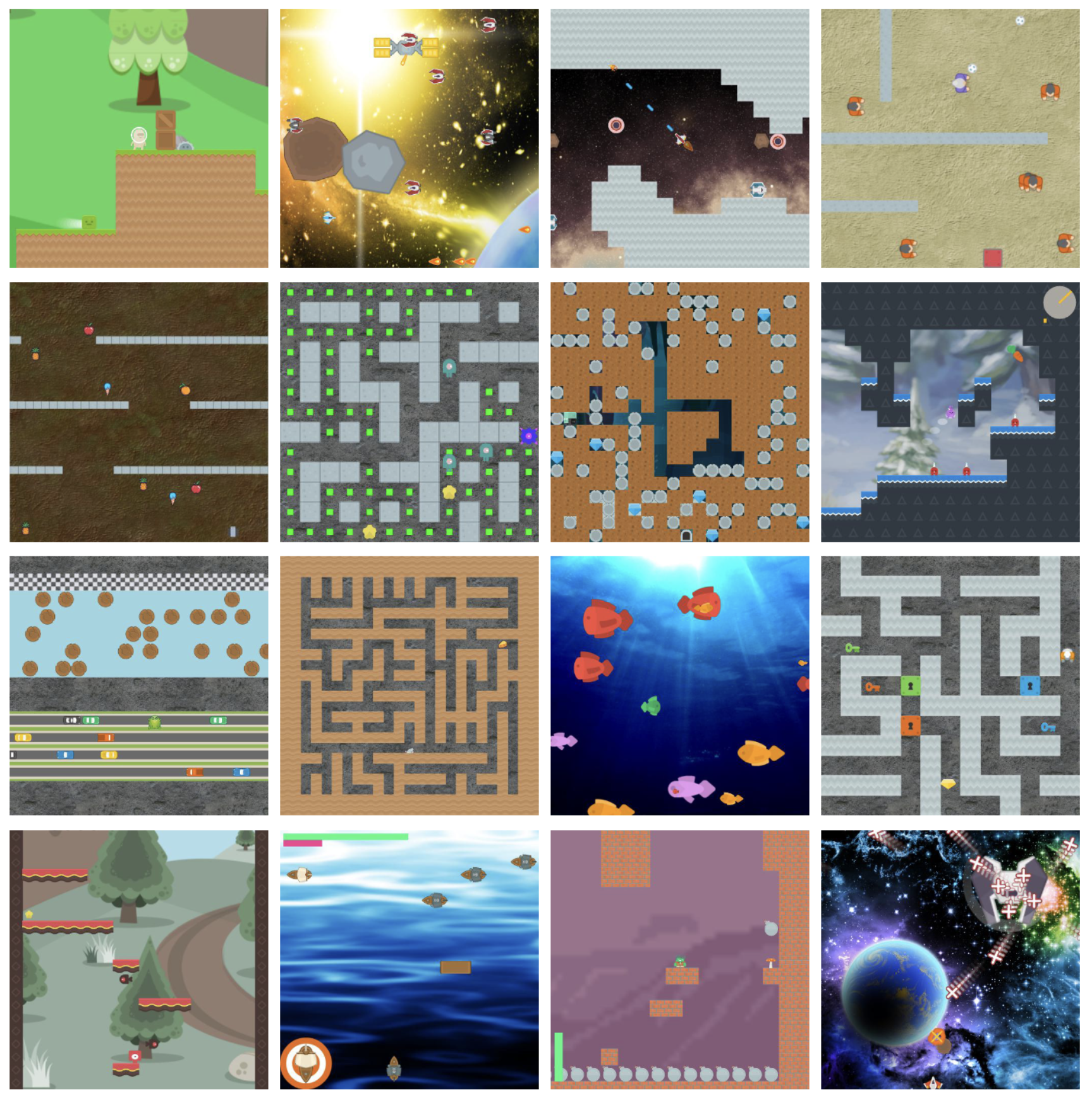}
    \caption{Screenshots of games in Procgen Benchmark \cite{pmlr-v119-cobbe20a}}
    \label{fig:procgen_sample}
\end{figure}

We used the \textit{Procgen} environment developed by OpenAI, a suite of 16 procedurally generated environments that allow the creation of adaptive environments with the use of different seeds \cite{pmlr-v119-cobbe20a}. The inherent mechanics of these environments serve as an ideal platform for evaluating a student model's ability to learn and adapt to variations introduced by different seeds. Furthermore, the diverse environments help our study as the agent is trained for a variety of challenges that may not occur during standard training procedures. This way we are allowed to scrutinize the agents' performance across different scenarios that could arise in practical implementations. Some sample games from the Procgen environment are shown in Figure \ref{fig:procgen_sample}. 

In Procgen environments, the \textbf{state space}, consists of 64x64 pixel images (RGB array with three channels and values $0$ to $255$). The \textbf{action space} is discrete in nature and usually includes movements (up, down, left, right) and interactions like collecting items or opening doors. For our experiments, we consider three procedurally generated games: Bigfish, Starpilot, and Jumper. In Starpilot for example, the player must navigate a space ship to avoid being hit by bullets and shoot down enemies in an arcade game fashion. Enemies and obstacles are procedurally generated, so each `map' is different. The key characteristic of the Procgen benchmark is that, given a different \textbf{seed}, the player encounters a different `map' though the game rules are unchanged. \emph{A key challenge for AI agents lies in how well they can adapt to seeds that they have not seen before}.



\subsection{Model architectures and training}

\subsubsection{Model training}
There are two model architectures that we consider -- the expert policy model and the student policy model. The \textbf{expert model} is the neural network that we use as an expert policy, which is then used to produce the offline dataset $\mathcal{D}$. The \textbf{student model} is the neural network which we train on the offline data to produce a policy $\pi_\theta$. Recall that the goal of offline RL is to optimize the parameters $\theta$ in order to produce a good policy, as described in section \ref{sec:offline-rl}. 

The expert model is an agent with convolutional architecture found in IMPALA \cite{espeholt2018impala}, following the convention in Procgen paper \cite{pmlr-v119-cobbe20a}. 
We trained the expert policy on the environment using proximal policy optimization (PPO) \cite{schulman2017proximal}.
We used the PFRL package and followed one community-created pytorch implementation on GitHub \cite{JMLR:v22:20-376, Lerrytang}. We trained the expert model for 25 million steps on 200 seeds until it achieved a satisfying level of performance on the environment. Hence, we trained three expert models in total for each of the three environments. This IMPALA network has three convolutional blocks. The first convolutional block has the output channel 16, the next two blocks has the output channel 32. This setup has a total number of $9712 + 41632 + 41632 + 524544 + 3855 + 257 = 621632$ trainable parameters. Table \ref{tab:model-hyper} shows more details regarding expert model.

For the student (the model that tries to mimic the expert), we use CNN (convolutional neural network) as our base model. The CNN model has 4 convolutional modules followed by a fully connected layer to the logits.  The convolutional layers utilize a $3 \times 3$ kernel with $3$ output channels and coupled with a ReLU activation and a average $2$ dimensional pooling layer with pool kernal size $2$ and stride $2$. For first $2$ layers, the dimension of output channel is $4$ times larger than that of the input channel. For the last $2$ layers, the dimension of output channel is $4$ times smaller than that of the input channel. So the output channel of the last layer is the same as the input channel of the first layer. The output of the convolutional layers is then passed to a fully connected layer, which maps to the logits. This leaner setup has a total number of $336 + 5232 + 336 + 327 + 735 = 6966$ trainable parameters, reducing the size of the fully connect layer. Student model has much fewer trainable parameters comparing to that of the expert model. Table \ref{tab:model-hyper} contains hyperparameters of training student models.

For each database collection method, we train 10 students and take the average of reward mean and reward standard deviation. We use Adam optimizer with learning rate 5e-3 \cite{kingma2017adam}. For behavioral cloning students, we train them with 1000 steps and batch size 256. For \method students, we train them with only 100 steps and batch size 15.

\begin{table}[hbt!]
    \centering
    \begin{tabular}{c|c c c}
    \toprule
         & Expert & BC Student & \method Student\\
    \midrule
         Model Params  & 621632 & 6966 & 6966\\
         Optimizer & Adam & Adam & Adam\\
         Learning Rate & 1e-5 & 5e-3 & 5e-3\\
         Batch Size & 8 & 256 & 15\\
         Steps & 25M & 1000 & 100\\
    \bottomrule
    \end{tabular}
    \caption{Hyperparameters: Expert V.S. BC Student V.S. \method Student }
    \label{tab:model-hyper}
\end{table}

\subsubsection{Data construction}
To construct offline RL dataset, we ran 100 episodes of our trained experts on all three environments, and store the data as mentioned in Section \ref{sec:offline-rl}.
To construct the synthetic data, given the synthetic data size, we sample data randomly from the offline RL data generated by expert, and then use gradient matching loss to udpate the synthetic data as illustrated in Figure \ref{fig:offline-distillation}. We use our experts to the RL Model to obtain gradients and compute the loss. Here, we use SGD optimizer with learning rate $0.1$ and momentum $0.5$, training with 1000 epochs.

\section{Results}

We compare our method to (1) the expert policy that we trained in an online fashion by interacting with the environment until we achieved good performance, which was then used to generate the offline dataset $\mathcal{D}$ and (2) a student trained on datasets with different levels of filtering using percentile behavioral cloning as described in section \ref{sec:behavioral_cloning}. In other words, we want to benchmark how well our method performs at \emph{generating a quality dataset that can be used for training an offline RL model}. 

\begin{figure}[htbp]
    \centering
    \begin{minipage}{0.48\textwidth}
        \centering
        \includegraphics[width=\linewidth]{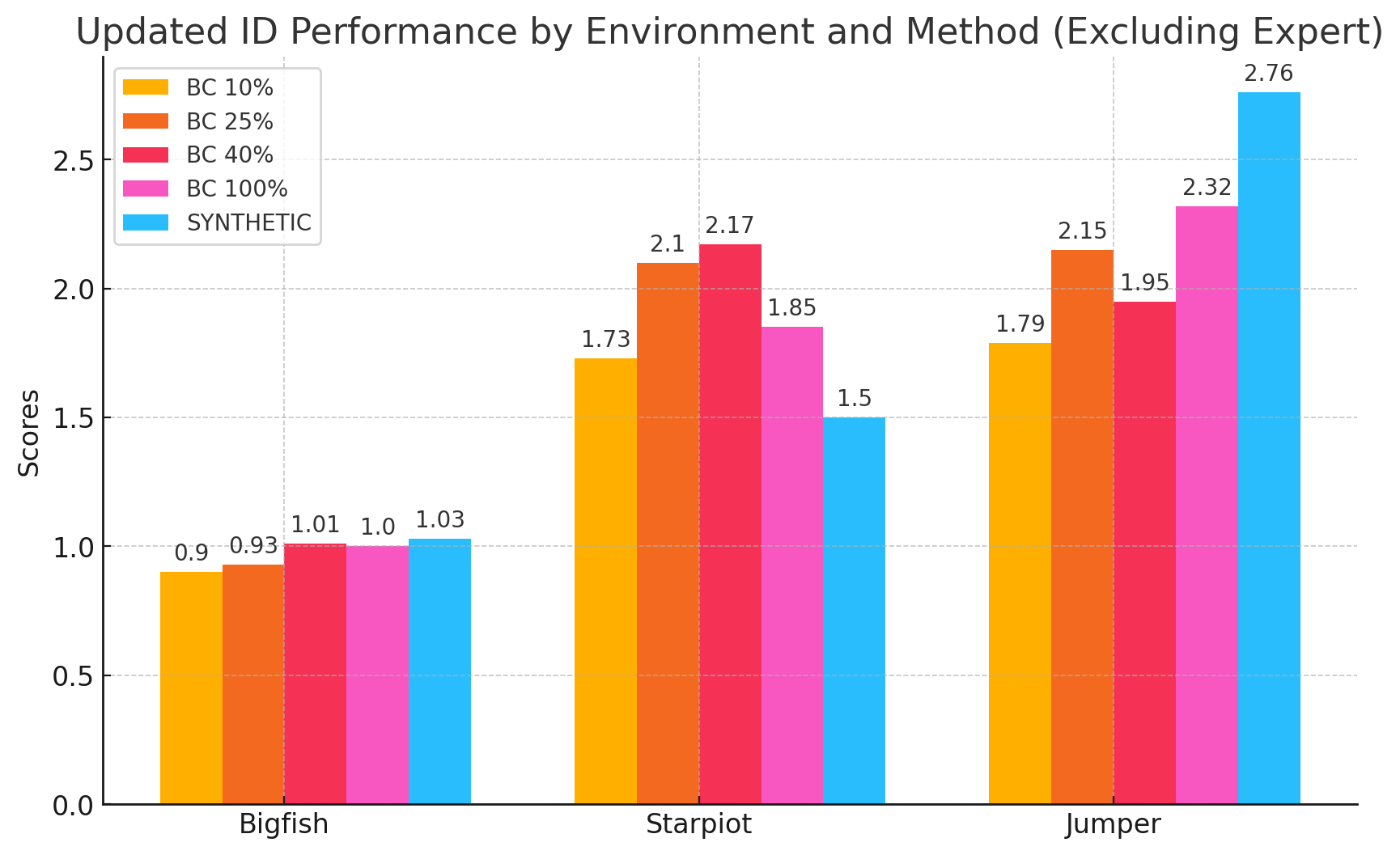}
        \caption{\textbf{In distribution performance of various data collection methods}}
        \label{fig:ID-performance}
    \end{minipage}\hfill
    \begin{minipage}{0.48\textwidth}
        \centering
        \includegraphics[width=\linewidth]{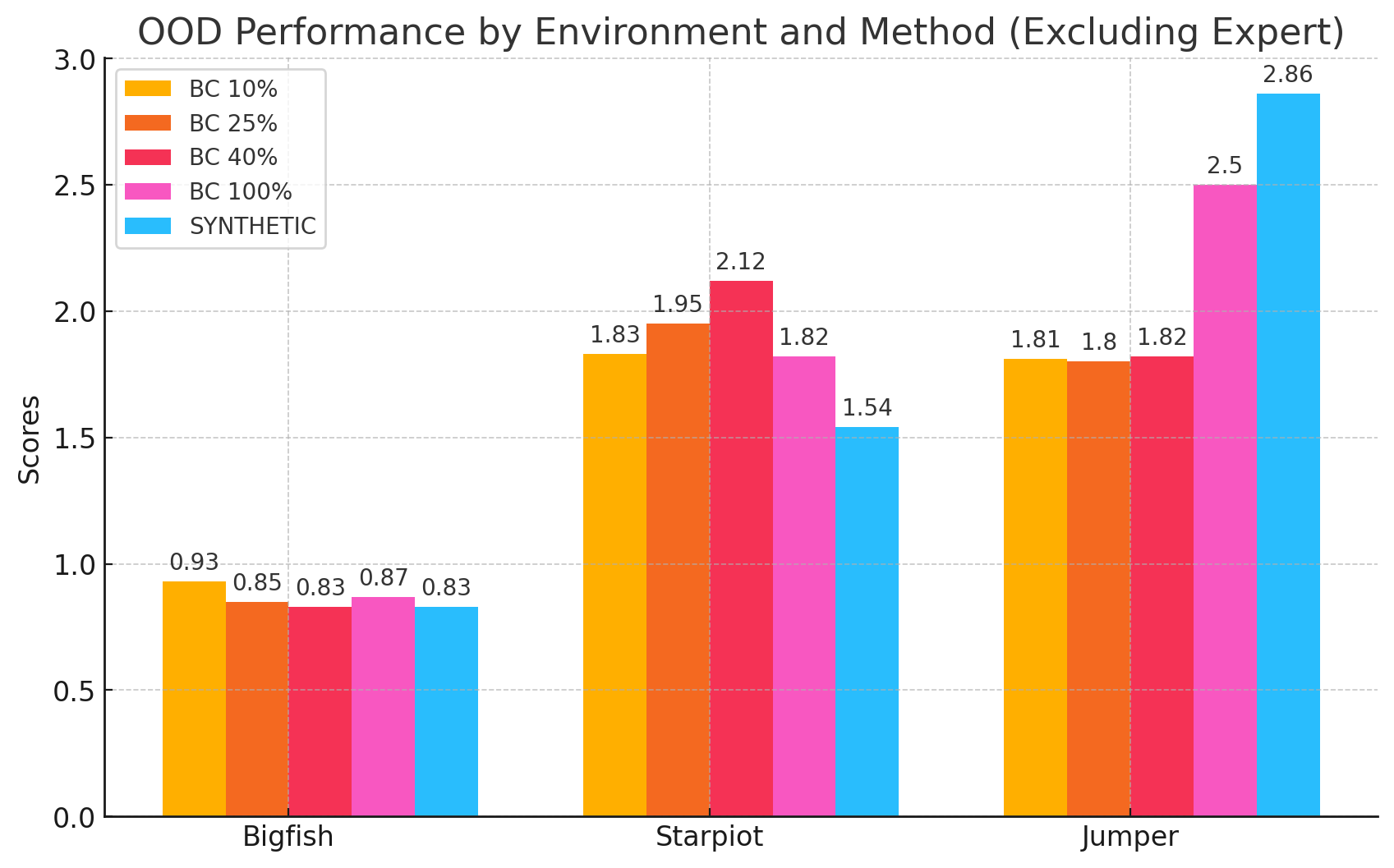}
        \caption{\textbf{Out of distribution performance of various data collection methods}}
        \label{fig:OOD-performance}
    \end{minipage}
\end{figure}

The student model was trained with the same number of stochastic gradient descent steps and same batch size for all baseline methods. We show the in-distribution performance, i.e. the performance on seeds contained in the offline dataset $\mathcal{D}$, of the various methods in table \ref{tab:dataset_id} and figure \ref{fig:ID-performance}. We see here that \method outperforms all percentile behavior cloning methods in both Jumper and Bigfish environments. \method does not outperform percentile BC on Starpilot. We observed that the Starpilot expert mainly takes one action during game -- the `shoot' action -- compared to Bigfish and Jumper where the expert takes a more even distribution of actions. Hence the dataset is hard to distill because of the imbalanced samples of different actions.  


Since we can procedurally generate out of distribution (OOD) scenarios in Procgen, we also tested the OOD performance of the various dataset generation methods, as shown in table \ref{tab:dataset_ood} and figure \ref{fig:OOD-performance}. We see here that similar, \method outperforms percentile behavior cloning methods in Jumper environment. \method also matches all percentile BC performances on Bigfish. \method does not outperform percentile BC on Starpilot since the dataset of Starpilot is imbalanced.

Our student model trained by \method only uses $150$ data samples, as shown in table \ref{tab:dataset_size}. Given \textbf{much smaller dataset size} (less than half of BC10\%) and \textbf{much fewer training steps} as mentioned in table \ref{tab:model-hyper}, \method achieves competitive results compared to behavioral cloning in different setups. \method also generalizes well out of distribution as the OOD performances matches the ID results, and often times in both Starpilot and Jumper outperforms the ID results.

\begin{table}[hbt!]
\centering
\begin{tabular}{c | c c c c c c} 
\toprule
\multicolumn{7}{c}{\large{ID Performance}}\\
\midrule
\textbf{Environment} & \textbf{Expert}  & \textbf{BC 10\%} & \textbf{BC 25 \%} & \textbf{BC 40 \%} & \textbf{BC 100 \%} & \textbf{\method} \\ 
\midrule
Bigfish & \smaller{14.27 $\pm$ 15.53} &  \smaller{0.90 $\pm$ 1.44} & \smaller{0.93 $\pm$ 1.47}  & \smaller{1.01 $\pm$ 1.62} & \smaller{1.00 $\pm$ 1.67} & \textbf{ \smaller{1.03 $\pm$ 1.99}} \\
Starpilot & \smaller{28.88 $\pm$ 19.41} & \smaller{1.73 $\pm $ 2.20} & \smaller{2.10 $\pm$ 2.66} & \smaller{\textbf{2.17 $\pm $ 2.58}} & \smaller{1.85 $\pm$ 2.22}& \smaller{1.5 $\pm$ 1.96}\\ 
Jumper & \smaller{8.79 $\pm$ 3.26} & \smaller{1.79 $\pm$ 3.54} & \smaller{2.15 $\pm$ 4.12} & \smaller{1.95 $\pm$ 3.86} & \smaller{2.32 $\pm$ 4.13} & \smaller{\textbf{2.76 $\pm$ 4.40}}\\
\bottomrule
\end{tabular}
\caption{\textbf{Average in distribution performance of student trained on various data collection methods.}}
\label{tab:dataset_id}
\end{table}

\begin{table}[hbt!]
\centering
\begin{tabular}{c | c c c c c c} 
\toprule
\multicolumn{7}{c}{\large{OOD Performance}}\\
\midrule
\textbf{Environment} & \textbf{Expert}  & \textbf{BC 10\%} & \textbf{BC 25 \%} & \textbf{BC 40 \%} & \textbf{BC 100 \%} & \textbf{\method} \\ 
\midrule
Bigfish & \smaller{6.03 $\pm$ 9.84}&\textbf{ \smaller{0.93 $\pm$ 1.38}} & \smaller{0.85 $\pm$ 1.19} & \smaller{0.83 $\pm$ 1.28} & \smaller{0.87 $\pm$ 1.32}&  \smaller{0.83 $\pm$ 1.04}\\
Starpilot & \smaller{23.34 $\pm$ 18.30}  & \smaller{1.83 $\pm$ 2.39} & \smaller{1.95 $\pm$ 2.39} & \textbf{\smaller{2.12 $\pm$ 2.35}} & \smaller{1.82 $\pm$ 2.20}& \smaller{1.54 $\pm$ 1.93} \\ 
Jumper & \smaller{5.61 $\pm$ 4.96}& \smaller{1.81 $\pm$ 3.51} & \smaller{1.8 $\pm$ 3.73}  & \smaller{1.82 $\pm$ 3.70} & \smaller{2.50 $\pm$ 4.26} & \smaller{\textbf{2.86 $\pm$ 4.43}}\\
\bottomrule
\end{tabular}
\caption{\textbf{Average out of distribution performance of student trained on various data collection methods.}  }
\label{tab:dataset_ood}
\end{table}

\begin{table}[hbt!]
\centering
\begin{tabular}{c | c c c c c} 
\toprule
\multicolumn{6}{c}{\large{Dataset Size}}\\
\midrule
\textbf{Environment}  & \textbf{BC 10\%} & \textbf{BC 25 \%} & \textbf{BC 40 \%} & \textbf{BC 100 \%} & \textbf{\method} \\ 
\midrule
Bigfish & \smaller{2027} & \smaller{5014} & \smaller{7336} & \smaller{10450}& \smaller{\textbf{150}}  \\
Starpilot & \smaller{1116} & \smaller{2796} & \smaller{4192} & \smaller{6830}& \smaller{\textbf{150}}\\ 
Jumper & \smaller{392} & \smaller{919}  & \smaller{1337} & \smaller{4837} & \smaller{\textbf{150}}\\
\bottomrule
\end{tabular}
\caption{\textbf{Dataset size used on various data collection methods with respect to different environments.}  }
\label{tab:dataset_size}
\end{table}

\section{Related work}

\subsection{Deep Reinforcement Learning}
Deep reinforcement learning has seen incredible success recently in tackling wide-ranging problems, from chess and Go to Atari games and robotics \cite{alphago}. We have also seen great improvements in the architecture used to design such agents, from PPO \cite{schulman2017proximal} to decision transformers \cite{chen2021decision}. As deep neural networks have shown their effectiveness in various tasks, researchers in reinforcement learning have increasingly turned their attention to them. Many complex reinforcement learning situations require these versatile neural networks for tasks such as encoding the states of the agents, learning complex policies, and assessing their performance. \cite{arulkumaran2017deep} give a good overview of various ways deep neural networks were incorporated into reinforcement learning settings.

\subsection{Knowledge Distillation}
As using large deep neural networks started bringing remarkable success in multiple real-world scenarios related to large-scale data, it became important to deploy deep models within mobile devices and embedded systems. \cite{bucilua2006model} first addressed this issue and proposed compression of large models for transferring the information from large models to train a small model such that accuracy is not hampered. \cite{hinton2015distilling} popularized the term 'knowledge distillation' as the process of learning a small model from a large model (teacher) to a small student model.  
In recent times, there have been many extensions to knowledge distillation where the focus is on compressing deep neural networks. The lightweight student models have paved the way for integrating knowledge distillation in various applications like adversarial attacks \cite{papernot2016distillation}, security and privacy of data \cite{wang2019private}, data augmentation \cite{lee2020self} etc. KD has been a key instrument in the study of natural language processing (NLP) \cite{devlin2018bert}. \cite{sun2019patient} and \cite{tang2019distilling} have used some lightweight variations of BERT (called BERT model compression) through knowledge distillation. \cite{jiao2019tinybert} proposed a TinyBERT, a two-stage transformer knowledge distillation, to make the framework even lighter. 

\subsection{Policy distillation}

There have also been attempts to distill the policy of a expert (teacher) network down to a student network directly \cite{rusu2015policy}, known as policy distillation. 
Policy distillation is a specialized application of knowledge distillation where it adapts the principles of KD in the context of Reinforcement Learning. It is used to transfer knowledge from one policy to another in deep RL. \cite{czarnecki2019distilling}identified three techniques for distillation in DRL, making comparisons of their motivations and strengths through theoretical and empirical analysis, including expected entropy regularized distillation, which ensures convergence while learning quickly. Policy distillation can also be used to extract an RL agent's policy to train a more efficient and smaller network that performs expertly \cite{rusu2015policy}.


While our work also teaches a student a policy based on some teacher policy, we take more indirect, offline approach where the student is only allowed to see offline data generated by the expert (teacher). 


\subsection{Dataset Distillation}
Dataset distillation is a dataset reduction method that synthesizes a smaller. 
In the original work, this is by feeding a randomly initialized model with samples from the real data and samples from the synthetic dataset and taking the gradient of the model with respect to these two data samples \cite{wang2018dataset}. The difference between the two gradients is taken as the loss, and the values of the data in the synthetic dataset are updated using SGD (while keeping the model weights fixed)
Since then, a wide variety of different distillation methods have been proposed (\cite{yu2023dataset}, \cite{lei2023comprehensive}). 
In one such work, instead of matching the gradients for a single sample, the sum of the gradients (total change in model parameters) after training on a series of samples is matched instead (\cite{cazenavette2022dataset}).
Despite recent interest in this technique, to the best of the author's knowledge, there have not been any applications of dataset distillation to reinforcement learning yet. 

\subsection{Task Generalization}
The goal of task generalization is to transfer what is learned from doing one task to other tasks, broadening the capabilities of the model. In the ideal scenario, the learned model should be able to apply its knowledge to changing tasks by using the core knowledge learned. (\cite{taylor2007cross}) suggests a new transfer method called "Rule Transfer" which aims to learn the rules of a source task and apply them to other target tasks. (\cite{taylor2008autonomous}) aims to learn mappings between the transitions from the source to the target task. In the problem suggested in (\cite{oh2017zero}), agents are required to learn to execute sequences of instructions after mastering subtask-solving skills. The problem gives out a good basis for generalizing to unseen tasks. In (\cite{lehnert2020reward}), authors suggest that using reward predictions gives the agents better generalization capabilities.

\subsection{Other Works}
In the work (\cite{yuan2021reinforced}), the authors propose a Reinforcement Learning based method for Knowledge Distillation for scenarios where multiple teachers are available. Their work is focused on NLPs and uses large pre-trained models like BERT and RoBERTa, where the framework dynamically assigns weights to different teacher models on each training instance, in order to maximize the performance of the distilled student model. 
In  (\cite{xie2021explore}), the authors present a novel framework (DRL-Rec) for knowledge distillation between RL-based models in list-wise recommendation, where they introduce one module by which the teacher decides on which lesson to teach to the student and another confidence-guided distillation through which it is determined how much the student should learn from each lesson.

\section{Conclusion and limitations}


There are several limitations to our study. The first one is that, limited by the computation resources and time we had, we only tested on three environments in Procgen. However, we believe that the experiments on these environments demonstrate the potential of our method, and we look forward to future work on other environments. 
We also focus on imitation policy learning in our work since our emphasis is on the dataset distillation, not the policy learning method. However, it is also possible to use other RL methods such as q-learning or actor-critic to train policies on the synthetic dataset. 
We mainly benchmark against percentile behavior cloning, since that is that is the closest method in the existing literature that `filters' for a better quality training dataset. 

\textbf{In conclusion,} we proposed and tested a method that synthesizes a better quality training dataset for offline reinforcement learning. The performance of our method suggests that the quality of the dataset a key component to training a better model, and a smaller but higher quality dataset can lead to similar or better performance compared to a larger one. 
We believe that these methods can be highly effective in settings with low amounts of data or noisy data, and where data cannot be collected online. 
There is still much to explore in this research space. 

\bibliography{dmlr_main}
\end{document}